\documentclass[letterpaper, 10 pt, journal, twoside]{IEEEtran}
\usepackage{floatrow}
\newfloatcommand{capbtabbox}{table}[][\FBwidth]
\usepackage{graphicx}
\usepackage{comment}
\usepackage{amsmath,amssymb} 
\usepackage{color}
\usepackage{hyperref}
\usepackage{subfigure}

\newcommand{\tabincell}[2]{\begin{tabular}{@{}#1@{}}#2\end{tabular}} 

\begin{document}
%
\title{Co-Planar Parametrization for Stereo-SLAM and Visual-Inertial Odometry}

%
%
%

\author{Xin Li$^{1,*}$, Yanyan Li$^{2,*}$, Evin P{\i}nar \"{O}rnek$^{2}$, Jinlong Lin$^{1}$ and Federico Tombari$^{2,3}$
\thanks{Manuscript received: May, 12, 2020; Revised August, 6, 2020; Accepted September, 22, 2020.}
\thanks{This paper was recommended for publication by Editor Sven Behnke upon evaluation of the Associate Editor and Reviewers' comments.} 
\thanks{$^{1}$Xin Li and Jinlong Lin are with Peking University, China}%
\thanks{$^{2}$Yanyan Li, Evin P{\i}nar \"{O}rnek$^{2}$ and Federico Tombari are with Department of Informatic, Technical University of Munich, Germany}
\thanks{$^3$:Google Inc.}
\thanks{$*$:The first two authors contributed equally. The authors would like to thank Dr. Yijia He for the helpful discussion.}
\thanks{Digital Object Identifier (DOI): see top of this page.}
}

%
%

\markboth{IEEE Robotics and Automation Letters. Preprint Version. Accepted September, 2020}
{Li \MakeLowercase{\textit{et al.}}: Co-Planar Parametrization} 

%



\maketitle

\begin{abstract}
This work proposes a novel SLAM framework for stereo and visual inertial odometry estimation. It builds an efficient and robust parametrization of co-planar points and lines which leverages specific geometric constraints to improve camera pose optimization in terms of both efficiency and accuracy. 
The pipeline consists of extracting 2D points and lines, predicting planar regions and filtering the outliers via RANSAC. Our parametrization scheme then represents co-planar points and lines as their 2D image coordinates and parameters of planes. We demonstrate the effectiveness of the proposed method by comparing it to traditional parametrizations in a novel Monte-Carlo simulation set. Further, the whole stereo SLAM and VIO system is compared with state-of-the-art methods on the public real-world dataset EuRoC. Our method shows better results in terms of accuracy and efficiency than the state-of-the-art. The code is released at \url{https://github.com/LiXin97/Co-Planar-Parametrization}.
\end{abstract}


%
\IEEEpeerreviewmaketitle

\section{INTRODUCTION}
\label{sec:intro}

Simultaneous Localization and Mapping (SLAM) and Visual Inertial Odometry (VIO) algorithms aim at camera pose estimation and scene reconstruction under unknown environments. They are ubiquitously employed in robotics for tasks such as planning, obstacle avoidance and navigation. When applied to indoor environments, these methods have to face important challenges due to the poor visual features available in the scene, which is often mostly characterized by low textured surfaces. 

It has been shown that the structural regularities in the environment (\textit{e.g.} lines and planes) bring valuable information to both SLAM and VIO systems~\cite{lu2015visual, he2018pl}. Such features can guide the SLAM optimization process by introducing additional constraints. However, how to organize such structural information and integrate it with the optimization in an efficient way is still an open question. Traditional representations focused on improving the trajectory accuracy, yet they ignored the high computational burden. In this work, we aim to tackle this problem by designing a better representation for planar structures, which simultaneously improves the accuracy and the efficiency of integrated stereo SLAM and VIO systems. 


So far in the literature, several works leveraged points and lines detected from an RGB image to handle challenging environments~\cite{he2018pl, mur2015orb, engel2017direct, qin2018vins}. Yet, the inner geometric relationship between those features is ignored in most of them. Different than using independent features of line segments and points, planar regions require fewer parameters to represent environments. Such planar regions and features can be found in almost all man-made environments, and they have been studied and leveraged in stereo SLAM and VIO systems~\cite{lu2015visual, rosinol2019incremental, zou2019structvio, li2020leveraging, Rosinol19arxiv-Kimera}. They introduce more constraints to the system that are helpful to improve overall accuracy. Nevertheless, they also rely on a high number of optimization parameters yielding limitations in real-world scenarios.

\begin{figure}[t]
    \centering
    \includegraphics[scale=0.36]{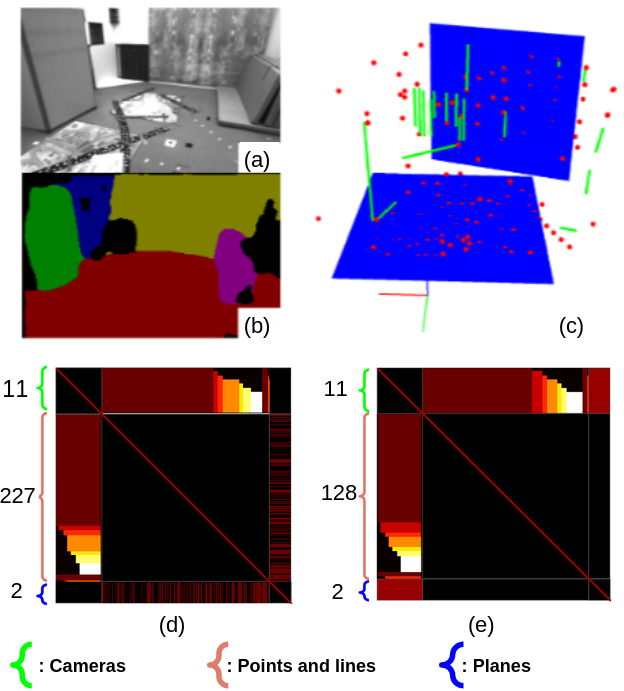}
    \caption{System results: (a) input RGB frame; (b) plane instance segmentation; (c) reconstruction for points, lines and infinite planes; (d) and (e) Hessian matrices that show the spatial correlation of camera and 3D landmarks within the traditional~\cite{civera2008inverse, bartoli20013d} and proposed parametrizations, respectively. Black areas represent zeros, non-zeros otherwise. (e) is sparser than (d). Number of camera parameters (green), points and lines features (orange) and plane parameters (blue) are shown.}
    \label{fig:parametrization1}
\end{figure}

In this work, we propose a novel method to employ planarity constraints to improve the accuracy and efficiency of SLAM models based on VIO or stereo in indoor environments. Our method detects the co-planar point and line features through a deep learning based plane detection followed by RANSAC filtering. We then introduce a novel parametrization to represent these co-planar features in an unified manner instead of using them as independent features. 
The resulting parametrization decrease the size of Hessian matrix, as well as make it sparser as shown in Fig.~\ref{fig:parametrization1}(e). As a result, solving the bundle adjustment problem for estimating the correct camera parameters and 3D landmarks through second-order Newton optimization, which relies on calculating Schur complement on the Hessian matrix, becomes more efficient.

\begin{figure*}[t]
\vspace{5pt}
    \centering
    \includegraphics[scale=0.29]{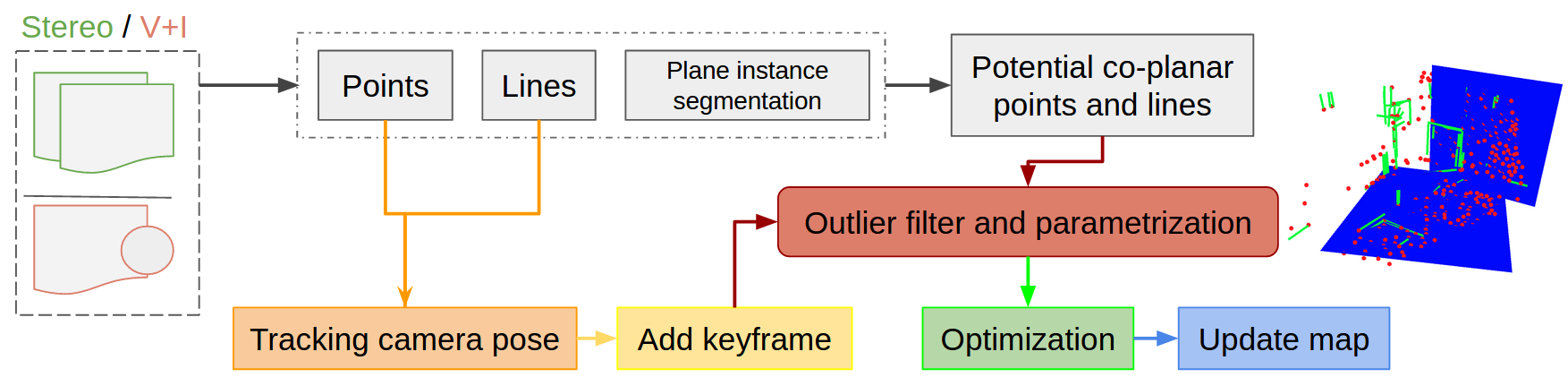}
    \caption{The pipeline of our plane-parametrized SLAM system. The overall pipeline follows the classical tracking and mapping approaches~\cite{mur2015orb}, along with the sliding-window based optimization. The pipeline can take as input either a stereo image pair or an image with IMU sensor data. 2D features and initial camera pose is estimated in a similar way as previous works~\cite{mur2017orb, qin2018vins}. Then we detect planar regions via plane instance segmentation. After selecting the potential co-planar points and lines on the planar region, we remove the outliers with RANSAC. We present the remaining robust points and lines with the proposed parametrization, which can be directly integrated as an additional constraint in SLAM optimization.}
    \label{fig:pipeline}
\end{figure*}

Furthermore, we show how our parametrization model can be integrated in a stereo SLAM or VIO pipeline as shown in Fig.~\ref{fig:pipeline} as we want to prove that our plane extraction and parametrization methods are general. By taking either a stereo image input, or an image with IMU sensor data, we solve the tracking and mapping problem through a graph based optimization. The non-planar 3D landmarks are integrated in the traditional way as 3D points, whereas the planar landmarks are introduced within the pipelines through proposed co-planarity parameters.

For evaluation, we used the public real-world EuRoC dataset and a newly created Monte-Carlo simulation set to perform further ablation studies. We compare our stereo-SLAM method against point-line SLAM approaches, as well as our VIO method against the state-of-the-art plane-based VIO models. Our method shows improvement in accuracy on both pipelines while benefiting from lower runtime, demonstrating the effectiveness of co-planar constraints for SLAM. In summary, our paper proposes the following contributions:
\begin{itemize}
    \item a novel two-stage plane detection strategy from RGB images, leveraging a neural network based plane segmentation and a robust outlier filtering
    \item a novel parametrization for co-planar points and lines that unifies the parameters, resulting in an efficient bundle adjustment optimization through the smaller and sparser Hessian matrix 
    \item the deployment of these contributions within two different camera tracking frameworks, based respectively on VIO and stereo SLAM, both individually reporting state-of-the-art results.  
\end{itemize} 

\section{RELATED WORK}\label{sec:literature}

Feature-based SLAM is traditionally addressed by tracking keypoints along successive frames and then minimizing some error functions (typically based on re-projection errors) to estimate the camera poses \cite{cadena2016past}. For point/based only method, there are many successful proposals, such as PTAM~\cite{klein2007parallel}, SVO~\cite{forster2016svo} and ORB-SLAM~\cite{mur2015orb}. However, using only point features has strong limitations within textureless environments as well as under illumination changes. 

To deal with these problems, line-segment based methods were proposed~\cite{zhang2015building, gomez2019pl}. 
Moreover, planar regions and associated features have been leveraged by SLAM systems.
In early works~\cite{lu2015visual}, planes in the scene were detected by RANSAC among estimated 3D points, which is time consuming and not stable. 
These plane-based mapping and tracking methods, however, are common within RGB-D sensors since it is easier to segment planes from depth maps. Salas-Moreno~\textit{et al.}~\cite{salas2014dense} present a dense mapping approach by using bounded planes and surfels with RGB-D sensors. Point-Plane SLAM~\cite{zhang2019point} computes orthogonal relationships between planes from depth maps, then uses constraints for pose estimation. CPA-SLAM~\cite{ma2016cpa} models the scene as a global plane model, which is helpful to remove drift by aligning current RGB-D frame with the plane model. By using IMU, VIO methods can deal with fast motion easily.
MSCKF~\cite{mourikis2007multi} and ROVIO~\cite{bloesch2015robust} are popular filter-based methods, but the first one does not maintain estimates of 3D landmarks in the state vector. Different to those methods, an optimization strategy is used VINS-MONO~\cite{qin2018vins} and  Mesh-VIO~\cite{rosinol2019incremental} for pose estimation.

Instead of a set of features, planes are also used to construct co-planar regularities for points and lines. Instead of extracting planes from sparse point cloud, Mesh-VIO~\cite{rosinol2019incremental} builds 2D Delaunay triangulation based on 2D points first, and then project them into 3D from their correspondences. They find vertical and horizontal planes from the gravity vector given by the IMU, then merge the co-planar constraints in the optimization module. With the introduction of deep learning, methods were proposed to estimate planes from a single RGB image, hence opening up new possibilities for SLAM systems. PlaneReconstruction~\cite{YuZLZG19} and PlaneRCNN~\cite{Liu19CVPR} are state-of-the-art plane instance segmentation methods for a single image. In addition to planes, they also estimate depth and normal maps from a single RGB image.

%
Inverse depth~\cite{civera2008inverse} and parallax angle~\cite{zhao2011parallax} were proposed to represent point features in monocular systems. Inverse depth parametrization uses the inverse of the depth from its anchor camera, which works more accurately for distant features. Instead of using depth, the parallax angle is used in~\cite{zhao2011parallax} which obtains good performance in both nearby and distant features. TextSLAM~\cite{li2019textslam} suggests to extract text-based visual information and treats each detected text as a planar feature. In line parametrization methods,  Pl\"ucker coordinate is a popular representation method for 3D line initialization and transformation. Each 3D line, however, has only 4 degrees of freedom (4DoFs), and the six parameters of Pl\"ucker coordinates lead to over-parameterization~\cite{bartoli2005structure}. So, an orthogonal representation based on only four parameters is used in the optimization to solve this problem. 


\section{PROPOSED METHOD}\label{method}
In this section, we first explain our co-planar parametrization strategy, which includes plane instance detection and RANSAC based filtering steps. Then, we introduce the implementation details of our stereo and VIO versions that use the proposed parametrization in a sliding window optimization fashion.

\begin{figure}[t]
   \vspace{6pt}
    \centering
    \includegraphics[scale=0.5]{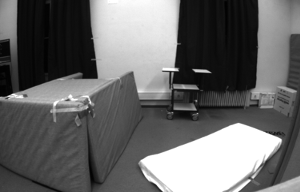}
    \includegraphics[scale=0.5]{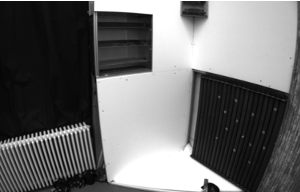}
    \includegraphics[scale=0.5]{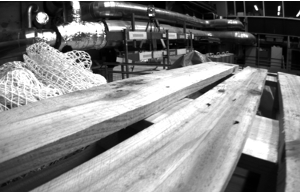}
    \includegraphics[scale=0.5]{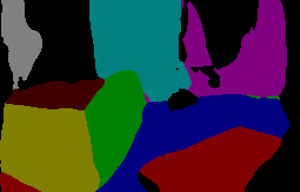}
    \includegraphics[scale=0.5]{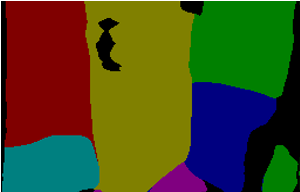}
    \includegraphics[scale=0.5]{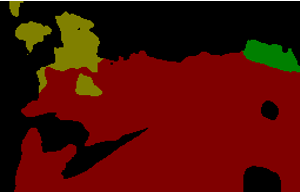}
    \includegraphics[scale=0.2]{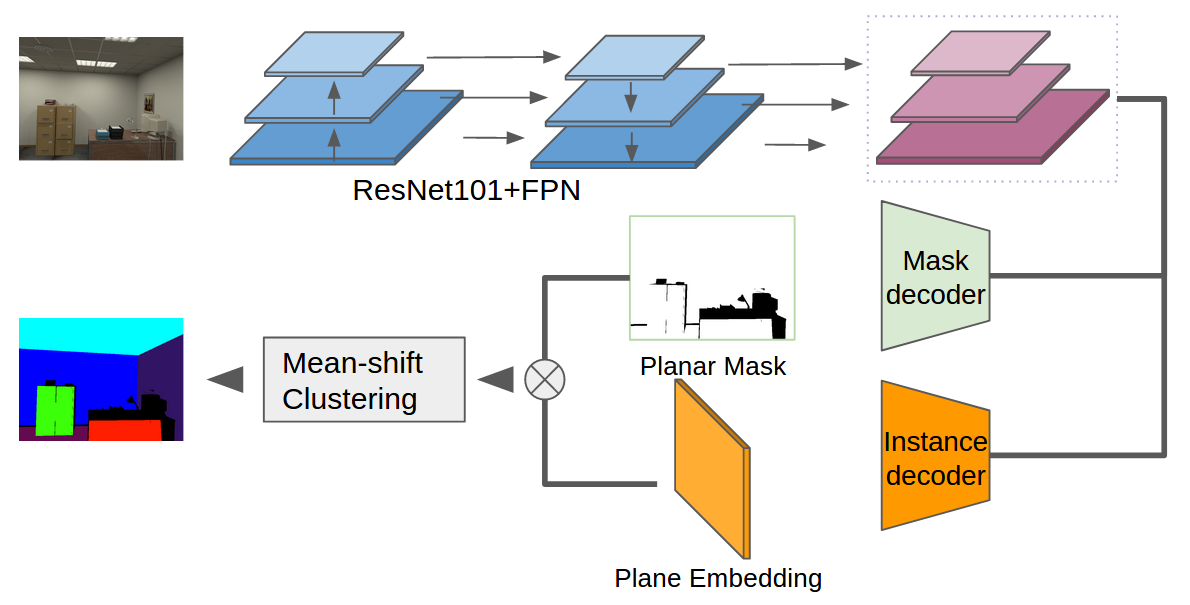}
    \caption{Examples of plane instance segmentation on EuRoC dataset and architecture of the plane instance segmentation network.}
    \label{fig:my_label}
\end{figure}

\subsection{Coplanarity-based Parametrization}\label{parametr}
A plane is defined by equation $\mathbf{n}X_c^T + d = 0$, where $\mathbf{n} = (n_1, n_2, n_3) \in R^3 $ is the normal of the plane, $X_c$ is a 3D point in camera coordinates, and $d \in R$ is the distance from the plane to the origin of the camera $c$. However, this representation has an over-parametrization problem, and it cannot be solved with the Gauss-Newton approach due to singularity issue ~\cite{kaess2015simultaneous}. So we optimize the normal $\mathbf{n}$ on the tangent space $S^2$ with another optimization method, which is similar to Mesh-VIO~\cite{rosinol2019incremental}. In this section, we first describe how the co-planar points and features are detected. Then, we explain our parametrization for points and lines, respectively.

\paragraph{\textbf{Plane Instance Segmentation}}
In order to detect planar regions in the scene in real-time, we use a plane instance segmentation network, which is a simplified version of PlaneReconstruction~ \cite{YuZLZG19}. This network has two branches: planar mask decoder and a plane embedding decoder. The first branch decodes a binary mask for planar regions. The second one decodes the feature maps to an embedding space where mean-shift clustering is used to group each pixel into planar instances, iteratively. We train this plane detection network on ScanNet dataset~\cite{dai2017scannet} for 30 epochs.

\paragraph{\textbf{Co-planar feature extraction}}
Since the plane instance segments extracted by the neural network might be at times inaccurate, we refine them by extracting 2D point and line features from images. Selecting the extracted features that align with the detected plane segments will lead us to robust features. We use ORB features~\cite{rublee2011orb} and LSD segment detection~\cite{von2012lsd} to extract sets of co-planar points $[S^x_1, \dots S^x_m]$ and co-planar lines $[S^l_1, \dots S^l_m]$, where each distinct set consists of co-planar features $S^x_n=[x_i \dots x_j], n\in[1,m]$ and $x_i$ is a 2D pixel.
For a stereo input, we obtain 3D points and lines by triangulating left-right image pairs. Whereas for VIO, the visual input is monocular and we triangulate sequential frames. During SLAM optimization, when a frame is detected as a new keyframe, we associate the features of this new frame with previous keyframes (i.e. check if they match and if they do not match, initiate new 3D landmarks with these features). After associating the landmarks, we build the potential co-planar points and lines, as shown in Fig.~\ref{fig:pipeline}. 

Due to the presence of outliers in the potential co-planar sets, we employ the following refinement strategy. First, for the current frame, we preserve the features that have been successfully triangulated. Then, we classify them according to detected 2D plane instance segments. If the number of features detected in a plane instance region is greater than a certain threshold, it will be considered as a potential planar region in 3D. If it is smaller than the threshold, plane will not be considered.
After that, we use a RANSAC filter to find co-planar constraints in the potential planar region. We take out points $C_x$ and lines $C_l$ in the potential planar region and feed them to the filter. Specifically, corresponding rules in Eq. \ref{eq:ransac} are selected to fit parameters $\Gamma$ of the plane according to the type of $z$ ($\forall z \in \mathcal{Z}, \mathcal{Z}=[C_x,C_l]$),
\begin{equation}
    f(c, \Gamma)=\left\{\begin{array}{l}
    \delta_{\perp}(P_x, \Gamma), z \in C_x \\
    max(\delta_{\perp}(c_{ls}, \Gamma), \delta_{\perp}(c_{le}), \Gamma), z \in C_l
    \end{array}\right.
    \label{eq:ransac}
\end{equation}
where $ \delta_{\perp}(\cdot, \cdot) $ denotes the perpendicular distance from a 3D point $P_x$ to the plane in 3D, $c_{ls}$ and $c_{le}$ are the start and end points of the line respectively. Note that we only consider lines which have both endpoints lie on the same planar region. 
If the size of the largest consensus set exceeds a threshold $\theta_{cp}$ (80\% in our experiments), we add the corresponding plane candidate to the system and establish point-plane and line-plane associations in the consensus set. We remove the outliers from the initial sets. When new 3D points and lines are generated in the system, we check if they belong to existing planes using the same metric defined above and store those correspondences.
It is important to note that it would be also possible to detect planar regions by using only RANSAC (without the deep learning method). However, when there are unknown number of planes in a scene, RANSAC does not work optimal. It requires several iterations, where at each time a single planar region is detected and inlier points are removed. Yet, the false-detections accumulate over each time and results degenerate. We prevent this issue by detecting all planes through a neural network initially.

\begin{figure}
\vspace{5pt}
    \centering
    \includegraphics[scale=0.45]{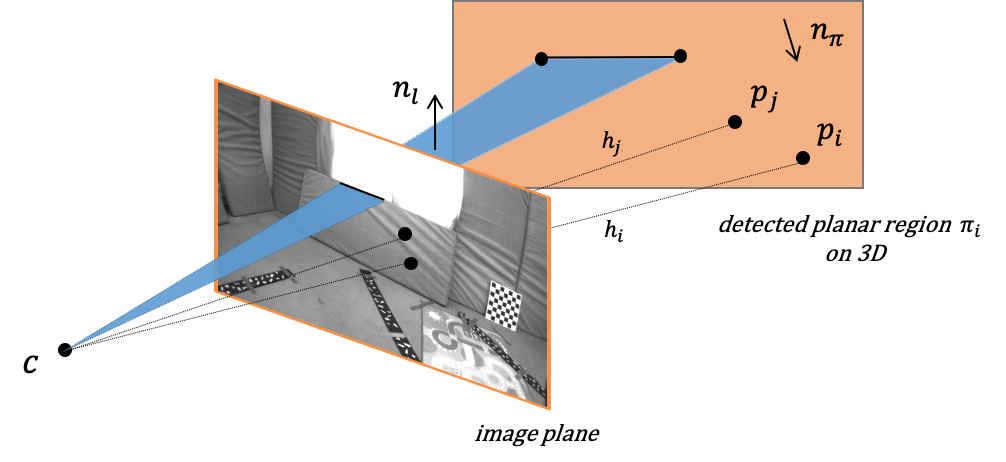}
    \caption{Point and line features are shown on a detected planar region $\pi_{j}$ with a normal $n_{\pi}$. $h_{i}$ is the depth from camera frame origin to the 3D point $p_i$. $n_{l}$ is the normal of a line on the plane $n_{\pi}$. Our parametrization rewrites the plane equation in terms of image pixel coordinates and combine line and point features. }
    \label{fig:parametrization}
\end{figure}

\paragraph{\textbf{Parametrization of points}}
After associating points and lines to co-planar regions as previously described, we obtain refined co-planar feature sets and parameters for each plane instance. As shown in Fig.~\ref{fig:parametrization}, 3D points are the intersections of the detected plane and the camera-to-landmark rays. For each 3D point $P^c_x=(x^c,y^c,z^c)$ which lies on the plane $\pi$ in camera frame $c$, we have the function $\mathbf{n}_{\mathrm{\pi}}^T P^c_x+d_{\pi}=0$. For example, the depth from origin of camera frame to the 3D point $P_i$ is $h_i$. A normalized 3D point is presented as $(\hat{x},\hat{y},1)$, where $(x^c,y^c,z^c) = (\hat{x},\hat{y},1)\cdot h_i$. Also,
\begin{equation}
    (\hat{x},\hat{y},1)^T = K^{-1}(u,v,1)^T 
\end{equation}
where $K$ is the intrinsic matrix of camera $c$, and $(u,v)$ is the 2D point corresponding to the landmark $P^c_x$. Then, the co-planar relationship for points can be represented as 
 \begin{equation}
     h_i\cdot \mathbf{n}_{\mathrm{\pi}}^T  K^{-1}(u,v,1)^T+d_{\pi}=0,
    \label{eq:point_para}
 \end{equation}
where the relationship contains 2D pixel of the landmark and parameters of the plane. So in our parametrization, the point $\mathbf{p}^*$ lying on a planar region can be represented as 
\begin{equation}
    \mathbf{p}^*= [\mathbf{n}_{\pi},d_{\pi}].
\end{equation}

\paragraph{\textbf{Parametrization of lines}}	
For line features, the Pl\"ucker coordinates  $\mathcal{L}=\left[\mathbf{n}_{\mathrm{l}}^{\top},\mathbf{d}^{\top}\right]^{\top} $ are used to initialize 3D lines, where $\mathbf{d} \in \mathbb{R}^3$ is the line's direction vector in camera frame $c$, and $\mathbf{n}_{\mathrm{l}} \in \mathbb{R}^3$ is the normal vector of the plane determined by the line and the camera frame's origin point (Fig.~\ref{fig:parametrization}). Furthermore, the line is the intersection of two known planes $\pi_l $ and $\pi_P $, so the dual Pl\"ucker matrix $\mathbf{L}^*$ can be computed by:	
\begin{equation}	
\mathbf{L}^{*}=\left[ \begin{array}{cc}{[\mathbf{d}]_{ \times}} & {\mathbf{n}_{\mathrm{l}}} \\ {-\mathbf{n}_{\mathrm{l}}^{\top}} & {0}\end{array}\right]=\pi_{l} \pi_{P}^{\top}-\pi_{P} \pi_{l}^{\top} \in \mathbb{R}^{4 \times 4} \\	
\label{eq:line_para}
\end{equation}	
where $\left[ \cdot \right]_{\times}$ is the skew-symmetric matrix of a three-dimensional vector, and $\bf{\pi}= [\mathbf{n}, d]$ is a 4D vector. Then we can easily get Pl\"ucker coordinates $\mathcal{L}=\left[\mathbf{n}_{\mathrm{l}}^{\top},\mathbf{d}^{\top}\right]^{\top} $ from the dual Pl\"ucker matrix.

\paragraph{\textbf{Resulting Hessian matrix}} Compared with other proposed representations, which treat points and lines as independent features, our method uses one plane parameter to represent all co-planar features. Novel parametrization is then used in the bundle adjustment, which is solved by a second-order Newton optimization method, the Levenberg-Marquardt algorithm. This relies on taking the gradients of the residuals with respect to parameters (3D landmarks and camera poses) and solving the normal equations. Hence, when there are less number of parameters, Hessian matrix will be smaller. When there are less dependencies between the parameters, the sparse structure of Hessian can be employed more efficiently through Schur complement. The resulting Hessian matrix is illustrated in Fig.~\ref{fig:parametrization1} and it's effects on efficiency are further shown in Experiments section, in Tab. \ref{table. time2}. The optimization equations are explained in next subsection. Further interested reader is referred to \cite{lourakis2009}.




\subsection{System Implementation}\label{systemimplementation}
In this section, implementation details are introduced for both versions of our approach, i.e. the stereo SLAM and VIO, respectively.    

\paragraph{\textbf{Tracking}}
The goal of the tracking module is to extract 2D features and estimate the camera pose for each frame. 
In the stereo version, we estimate camera pose via point and line features, where stereo keypoints are defined by three coordinates $x_s=(u_L,v_L,u_R)$, here $(u_L,v_L)$ are coordinates on the left image and $u_R$ is the horizontal coordinate for the corresponding matches in the right image. Similar to points, lines between two images are matched by Line Band Descriptor (LBD)~\cite{zhang2013efficient}. Furthermore, motion model is used to provide an initial pose that is refined by a frame-to-frame tracking strategy similar to ORB-SLAM~\cite{mur2015orb}. 
%
Instead, for the VIO version, the initialization strategy of IMU is similar to VINS-Mono~\cite{qin2018vins}, which relis on a loose coupling strategy to align IMU pre-integration with the visual-only part. Different than visual-only (stereo) branch, the initial pose for optimization in VIO is obtained from IMU pre-integration~\cite{he2018pl,qin2018vins} so that the visual part can be regarded as a purely monocular version. Monocular keypoints are defined by two coordinates $x_m=(u_L, v_L)$ which are triangulated from multiple views. 

In the system, we use different strategies for keyframe detection in stereo and VIO pipelines. For the former one, a new keyframe can be added only after at least 20 frames. Each keyframe tracks more than 40 points and 10\% of keypoints should be new keypoints compared to the nearest keyframe. However, for the latter one, we consider the average parallax (with rotation compensation) of tracked features between two keyframes, which should be more than 10 degrees (similar to VINS-Mono~\cite{qin2018vins}).

\paragraph{\textbf{Mapping}}
When a keyframe is detected and inserted, we associate its 2D features to 3D corresponding landmarks in the sliding window (or local map) by 2D feature matching. For each non-associated 2D feature, we triangulate it with other keyframes in the VIO version, while for stereo, non-associated points and lines are usually triangulated by each stereo pair. Different from points, 3D lines are triangulated by two intersecting planes colored in blue in Fig.~\ref{fig:parametrization}, which are observed in different views.

Based on the potential co-planar regions and the RANSAC filter, 3D landmarks are divided into two sets for optimization: planar features and non-planar features. Inverse depth algorithm is used to represent points; and Pl\"ucker coordinates and orthonormal representations are used to represent lines following He et al. ~\cite{he2018pl}, which are then fed to window-based bundle adjustment for optimizing poses and landmarks.

\paragraph{\textbf{Bundle adjustment with co-planar parametrization}}
In this part, we use re-projection error functions to optimize camera pose and landmark positions. Two different error functions are used for planar and non-planar features. Non-planar features are represented by traditional parametrization and optimized directly. However, co-planar features are refined by optimizing the parameters of the proposed parametrization.
For point features, the re-projection error $ \mathbf{r}^{\mathrm{p}}_{i k}$ strands for the the distance between the projected point of the $j$th map point and the observed point in the $k$th frame, which is noted as
\begin{equation} \label{error-point}
    \mathbf{r}^{\mathrm{p}}_{i k}= x_{ik}-\Pi(T_{kw},P_{i}^w)
\end{equation}
where $\Pi()$ re-projects the $i$th global 3D point $P_i^w$ coordinates into the $k$th frame. For general points, $P_i^w$ is represented as $(x^w,y^w,z^w)$. Points lying on a plane are represented with Eq.~\ref{eq:point_para}.

For line features, the re-projection error $\mathbf{r}_{jk}^{\mathrm{l}}$ is defined as the distance between the re-projected line of the $j$th map line and two endpoints of its corresponding 2D line in the $k$th keyframe, which is given by,
\begin{equation} \label{error-line}
 \mathbf{r}_{jk}^{\mathrm{l}} = \left[\begin{matrix}\frac{\mathbf{s}^{\top} \mathbf{n}_{\mathrm{l}} }{\sqrt{n_{1}^{2}+n_{2}^{2}}} && \frac{\mathbf{e}^{\top} \mathbf{n}_{\mathrm{l}}}{\sqrt{n_{1}^{2}+n_{2}^{2}}}  \end{matrix}\right]^\top   	
\end{equation}	
\noindent where $\mathbf{n}_{\mathrm{l}} = [ n_1, n_2, n_3 ]^\top$ is the 2D line re-projected from the 3D line to the camera frame, $\mathbf{s} = [\hat{x}_s,\hat{y}_s, 1]^{\top}$ and $\mathbf{e} = [\hat{x}_e,\hat{y}_e, 1]^{\top}$ are two end-points of the observed line segment in the $k$th image plane. For general lines, $\mathbf{n}_{\mathrm{l}}$ can be represented as in an orthonormal way~\cite{he2018pl}. Lines lying on a plane are represented with the Eq.~\ref{eq:line_para}.

Given by the Eq.~\ref{error-point} and Eq.~\ref{error-line}, We can therefore construct a unified target function which optimizes all terms simultaneously,
 \begin{equation} \label{eq:stereo_ba}
     E= \sum_{k,i} \rho_p({\mathbf{r}^{\mathrm{p}}_{i k}}^\top\Lambda_{ik} \mathbf{r}^{\mathrm{p}}_{i k}) + \sum_{k,j} \rho_l({\mathbf{r}_{jk}^{\mathrm{l}}}^\top\Lambda_{jk} \mathbf{r}_{jk}^{\mathrm{l}})
 \end{equation}
here $\rho_p$ and $\rho_l$ present robust Cauchy cost functions. Respectively, $\Lambda_{ik}$ and $\Lambda_{jk}$ are the information matrices of points and lines, as calculated in~\cite{he2018pl, qin2018vins}.

\paragraph{\textbf{Tightly-coupled optimization for inertial constraints}} For the VIO case, we fuse the data coming from the visual and inertial sensors via non-linear optimization in a tightly coupled form. Different from the stereo case, visual features are transferred to the IMU body coordinate system via extrinsic parameters $[\mathbf{R}_{bc} \quad \mathbf{t}_{bc}]$ between camera and IMU. So the unified target function for the VIO branch can be shown as,

 \begin{equation} \label{eq:vio_ba}
  \begin{aligned}
     E & = \sum_{k,i} \rho_p({\mathbf{r}^{\mathrm{p}}_{i k}}^\top\Lambda_{ik} \mathbf{r}^{\mathrm{p}}_{i k}) + \sum_{kj} \rho_l({\mathbf{r}_{jk}^{\mathrm{l}}}^\top\Lambda_{jk} \mathbf{r}_{jk}^{\mathrm{l}}) \\
     & + \sum_{b} \rho_l({\mathbf{r}^{\mathrm{b}}}^\top\Lambda_{b} \mathbf{r}^{\mathrm{b}}) + E_m
      \end{aligned}
 \end{equation}

where $\mathbf{r}^{\mathrm{b}}$ is the IMU residual, and $E_m$ is the prior residual from marginalization operator in the sliding window. For more details, readers are referred to~\cite{qin2018vins}.


\section{EXPERIMENTS}
\label{sec:experiments}

To evaluate the proposed method, we benchmark it against the state of the art on the EuRoC dataset~\cite{burri2016euroc}. In addition, we perform Monte-Carlo simulations to verify the robustness and efficiency of the novel parametrization. We evaluate both stereo and VIO pipelines with Absolute Trajectory Error (ATE) which measures absolute translational distances between the ground truth pose and the corresponding estimated pose. All the experiments run on an Intel Core i7-8550U @ 1.8GHz and 16GB RAM.

\subsection{EuRoC Dataset} \label{Experimental_intro}
	EuRoC is a popular public dataset for stereo SLAM and VIO systems, which collects stereo images and inertial data from an aerial vehicle in indoor environments~ \cite{burri2016euroc}. There are two scenarios in this dataset: Vicon Room (V) and Machine Hall (MH), with eleven sequences in total. VH is an indoor environment and has several planar regions, whereas MH is the interior of an industrial facility where planar regions are unevenly distributed. 

\paragraph{\textbf{Ablation studies}} In order to evaluate the performance of the proposed parametrization in EuRoC, we fix the front-end and compare five formulations: $P(-wo)$, $P(-w)$, $PL(-w)$, $P(-r)$, and $PL(-r)$, where $P$ denotes a point-based method, and $PL$ denotes a point-line-based system. $(-wo)$ means the traditional parametrization (only inverse depth), and both $(-r)$ and $(-w)$ use co-planar constraints in the optimization module, but in different ways. $(-r)$ uses more equations between point-to-plane and line-to-plane, which are merged into optimization as in Mesh-VIO~\cite{rosinol2019incremental, li2020leveraging}. Whereas $(-w)$ presents these residuals within the proposed co-planar parametrization. 


\begin{figure}
\vspace{5pt}
    \centering
    \subfigure[RMSE (cm), stereo]{\includegraphics[scale=0.233]{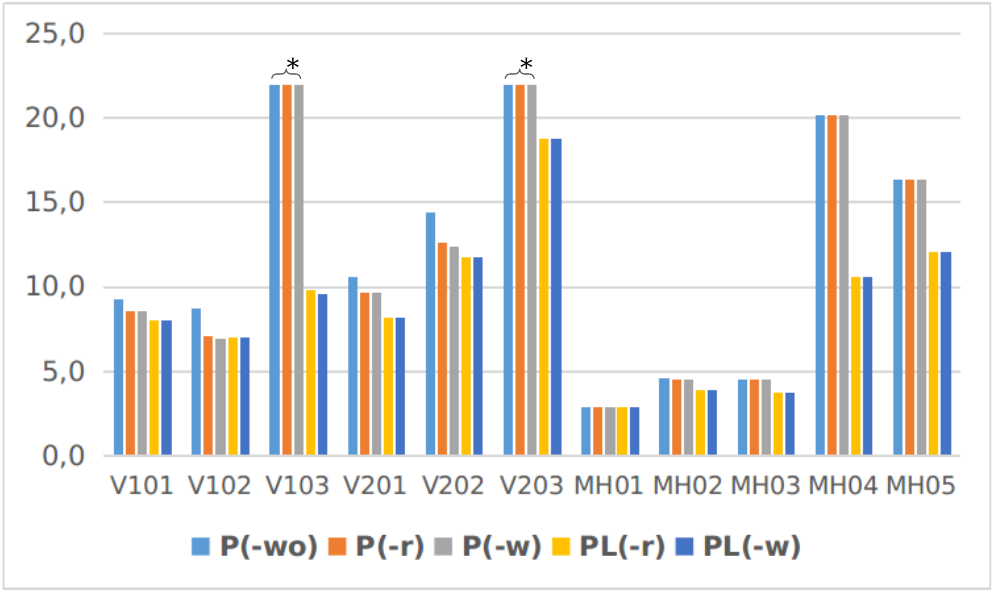}
    \label{fig:ATE_a}}
    \subfigure[RMSE (cm), VIO]{ \includegraphics[scale=0.64]{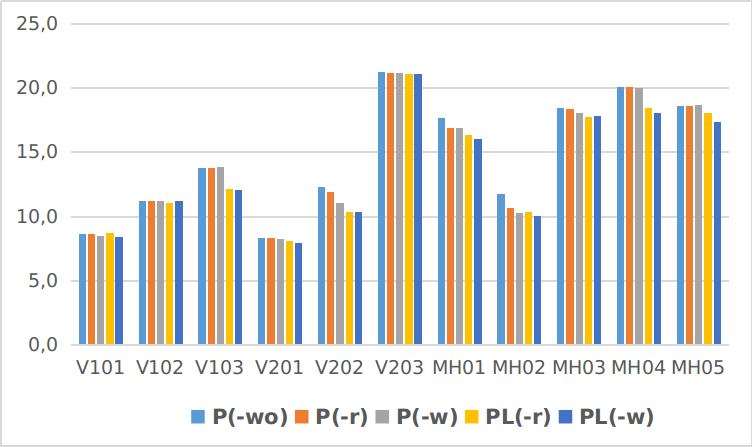}
    \label{fig:ATE_b}}
    \caption{Comparison in terms of ATE of different parametrization variants: $P(-wo)$, $P(-r)$, $P(-w)$,$PL(-r)$ and $PL(-w)$. The top part shows results for stereo, and the bottom one for VIO. The proposed parametrization $PL(-w)$ achieves the best results for all sequences where structural regularities are detected and enforced. * shows lost tracking on V103 and V203 sequences.}
\end{figure}    

The results of the stereo and VIO versions on EuRoC dataset are presented in Fig.~\ref{fig:ATE_a} and Fig.~\ref{fig:ATE_b}, respectively.      
In general, the proposed parametrization $PL(-w)$ results in lower RMSE compared to traditional parametrizations, $P(-wo)$ and $P(-w)$, in both cases, and especially in the MH sequences, where the line features can provide more robust constraints with planar regions in the large industrial environment.

For stereo approaches, as shown in Fig.~\ref{fig:ATE_a}, line features make the system more robust especially in $V103$ and $V203$ sequences, where severe motion blur happened. In other Vicon sequences, $PL(-w)$ performs better than $PL(-r)$ because the proposed two-stage co-planar approach removes distances between those co-planar features and planes directly. In $MH01$, $MH02$ and $MH03$ which are textured sequences, all approaches obtain similar results. 
In Fig.~\ref{fig:ATE_b}, $P(-wo)$ and $P(-w)$ perform equally on $MH03$, $MH04$ and $MH05$ sequences, because there are not any structural regularities detected. When there are some planar regions detected, as in $V202$, $MH01$ and $MH02$, the proposed parametrization $P(-w)$ obtains better performance than traditional methods. If enough features can be obtained and few good co-planar sets, our system's performance will degenerate to that of traditional methods, as in sequences $V102$ and $V201$. The computation time of different operations in V101 is presented in Tab.~\ref{table.real-time}.

\paragraph{\textbf{EuroC evaluation}}

We compare our stereo branch against the stereo version of ORB-SLAM2~\cite{mur2017orb} and FMD-SLAM~\cite{tang2019fmd}. It is important to note that, for fairness of comparison, the tested ORB-SLAM2 does not have loop closure. Furthermore, we compare our VIO version against the recently proposed MSCKF~\cite{mourikis2007multi}, ROVIO~\cite{bloesch2015robust}, VINS-MONO ~\cite{qin2018vins},
and Mesh-VIO~\cite{rosinol2019incremental}. Results are given in Tab.~\ref{table:ATE2}. These VIO algorithms use all a monocular camera, except Mesh-VIO that uses a stereo camera. Results of previous works are taken from Rosinol et al.~\cite{rosinol2019incremental}.

 \begin{table*}[htpb!]
 \vspace{8pt}
    \begin{floatrow}
    \scalebox{0.9}{
     \begin{tabular}{c|ccccccc|ccccc}
     \hline
    &\multicolumn{7}{c}{\textbf{VIO}}& \multicolumn{5}{|c}{\textbf{Stereo}} \\  
    &\tabincell{c}{MSCKF\\~\cite{mourikis2007multi}} &\tabincell{c}{ROVIO\\~\cite{bloesch2015robust}} &\tabincell{c}{VINS\\MONO~\cite{qin2018vins}} 
    &\tabincell{c}{Mesh \\VIO~\cite{rosinol2019incremental}} 
    &\tabincell{c}{PL\\(-wo)}
    &\tabincell{c}{PL\\(-r)}
    &\tabincell{c}{PL\\(-w)} 
    &\tabincell{c}{ORB\\SLAM2~\cite{mur2017orb}} &\tabincell{c}{FMD\\SLAM~\cite{tang2019fmd}} 
    &\tabincell{c}{PL\\(-wo)}
    &\tabincell{c}{PL\\(-r)}
    &\tabincell{c}{PL\\(-w)}   \\ \hline
     V101  &34 &10 &7 & \textbf{6} &8.4 &8.5 &8.4 
     &9 &9&8.4 &\textbf{8.0} &\textbf{8.0}
       \\ 
     V102  &20 &10 &10  &\textbf{7} &11.0 &10.9 &11.0 
     &8 &20 &7.5 &\textbf{7.0} &\textbf{7.0}
         \\ 
     V103  &67 &14 &13 &17 &\textbf{11.9} &\textbf{11.9} &\textbf{11.9} 
     &20 &53 &10.6 &9.8 &\textbf{9.6}
     \\ 
     V201  &10 &12 &\textbf{8} &\textbf{8} &8.1 &8.1& \textbf{8.0} &\textbf{7} &9 &9.0 &8.2 &8.2
     \\ 
     V202   &16 &14 &\textbf{8}  &10 &12.0 &10.5 & 10.5 
     &10 &\textbf{8}&12.4 &11.8 &11.8
     \\ 
     V203  &113 &\textbf{14} &21  &27 &20.9 &20.9  &20.9 
     &$\times$ &$\times$ &19.8 &\textbf{18.8} &\textbf{18.8}
     \\
    MH01  &42 &21 &27 &\textbf{14} &17.1 &16.3 &16.2 
    & 4 &4&\textbf{2.9} &\textbf{2.9} &\textbf{2.9}
     \\ 
    MH02  &45 &25 &12 &13 &11.0 &10.2 & \textbf{10.0} 
    &5 &4 &4.1 &\textbf{3.9} &\textbf{3.9}
     \\
    MH03  &23 &25 &\textbf{13} &21 &17.6 &17.6 &17.6 
    &4 &5 &4.0 &\textbf{3.7}&\textbf{3.7} 
    \\
    MH04  &37 &49 &23 &22 &18.5 &18.4 &\textbf{18.2} 
    &16 &\textbf{9}&10.6 &10.6 &10.6
     \\
    MH05  &48 &52 &35 &23 &18.2 &18.0 & \textbf{17.7} 
    &20 &\textbf{9}&12.1 &12.1 &12.1
    \\ \hline
    Average  &41.3 &22.3 &16.0 &15.2 &14.1 &13.8 &\textbf{13.7} &10.6* &12.9*&8.1* &\textbf{7.8*} &\textbf{7.8*}
    \\
    \hline
    \end{tabular}
    }
    \caption{Comparison in terms of RMSE (cm) of the proposed $PL(-w)$ pipeline against the state of the art on the EuRoC dataset. Best results are bolded. $\times$ shows lost tracking. Averaged results with * do not include the sequence V203. }
    \label{table:ATE2}
    \end{floatrow}
\end{table*}

The left part of Tab.~\ref{table:ATE2} shows that the $PL(-w)$ approach is an accurate and robust method compared with state-of-the-art VIO methods on sequences.
Compared with Mesh-VIO~\cite{rosinol2019incremental}, which also uses planar information to build co-planar regularities in the optimization process, our method performs better on most sequences, where Mesh-VIO obtained more vertical planes from 3D mesh due to using gravity during plane detection. When horizontal and vertical planes are difficult to detect as in $V103$ and some of the MH sequences, Mesh-VIO tends to degenerate easily so that it cannot build co-planar constraints. In sequence $MH05$, we observe a 26\% improvement compared to the second best performing algorithm (Mesh-VIO), and in sequence $V103$, a 15\% improvement and 35\% improvement compared to VINS-MONO and Mesh-VIO, respectively. It can be seen that the optimization methods of VINS-MONO, Mesh-VIO and $PL(-w)$ are more robust than the filter-based MSCKF. Meanwhile, our method is more robust for indoor environments that have lots of co-planar regularities.

The stereo SLAM comparison is shown on the right side of Tab.~\ref{table:ATE2}. Stereo ORB-SLAM2 obtains comparable results to ours on all sequences except V203 and MH04. In those textured sequences, this method tracks the features in a stable and accurate way. Instead, V203 is a difficult sequence because of the fast motion and the strong illumination changes, and tracking fails for both ORB-SLAM2 and FMD-SLAM. Benefiting from using point and line features, our method is instead more robust and can deal also with this sequence. The average RMSE values, for fairness computed without taking sequence V203 into account, show that our method obtains 25.7\% and 38.7\% improvements compared to ORB-SLAM2 and FMD-SLAM, respectively.  

\subsection{Simulation Dataset}
\begin{figure}[h]
    \centering
    \subfigure[Sequence a]{
    \includegraphics[scale=0.24]{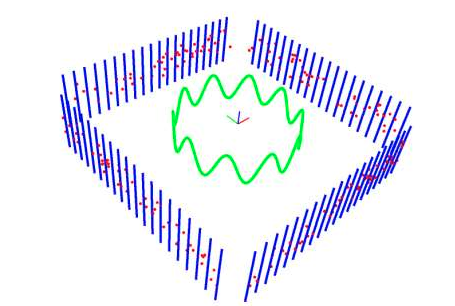}
    \label{fig:simulation_environment_a}}
    \subfigure[Sequence b]{
     \includegraphics[scale=0.24]{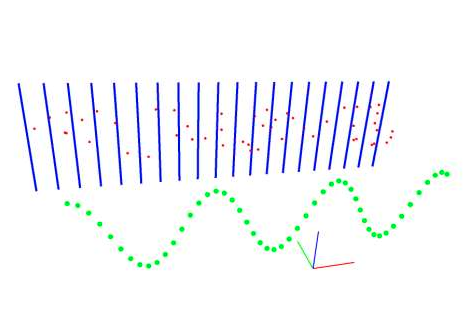}
     \label{fig:simulation_environment_b}}
    \caption{Two simulation environments are illustrated, where points and lines are in red and blue, respectively. Camera follows green trajectories.}
    \label{fig:simulation_environment}
\end{figure}

We create two simulation sequences with ideal co-planar environments to evaluate the efficiency with respect to performance under different parametric formulations. As shown in Fig.~\ref{fig:simulation_environment_a}, the first sequence has 100 lines and 200 points generated in 4 directions, which are observed by virtual cameras that follow a sinusoidal trajectory with 150 simulated poses. The second sequence consists of 20 lines and 50 points observed by 50 camera poses as shown in Fig.~\ref{fig:simulation_environment_b}. 

\begin{figure}[h]
\vspace{5pt}
  \includegraphics[scale=0.7]{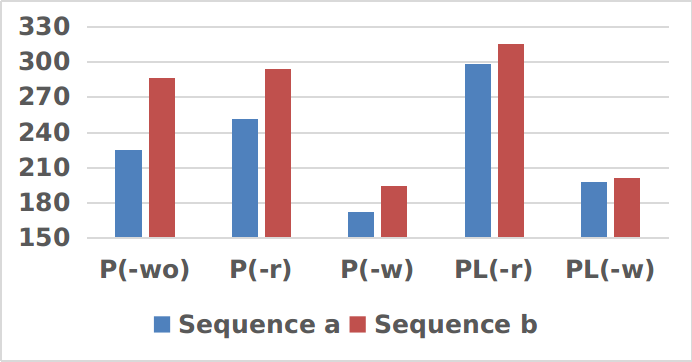}
  \includegraphics[scale=0.7]{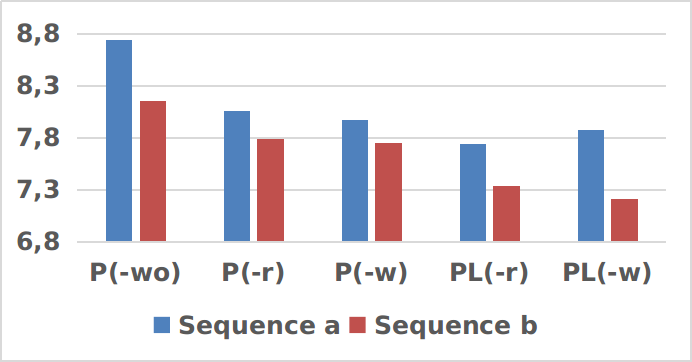}
    \caption{Comparison  of  the  optimization time (ms, top) and RMSE (cm, bottom) for pipelines $P(-wo)$, $P(-r)$, $P(-w)$, $PL(-r)$ and $PL(-w)$.}
    \label{fig:TIME_RMSE}
\end{figure}


For line measurements, the virtual camera gets two endpoints from each measurements. Note that each measurement of a point, including endpoints of lines and point features, is corrupted by 1-pixel Gaussian random noise. In order to simplify the simulation, we simulate relative pose odometry measurements as pose estimation results from the tracking module, which have random noise as,
\begin{equation}
    \bar{q}_{m}=\left[\begin{array}{c}
    \frac{1}{2} \mathbf{n}_{\theta} \\
    1
    \end{array}\right] \otimes \bar{q}, \quad \mathbf{p}_{C m}=\mathbf{p}_{C}+\mathbf{n}_{p}
\end{equation}
where $\mathbf{n}_{\theta}$ and $\mathbf{n}_p$ are the Gaussian white noises added to the relative pose, with $\sigma_{\theta}=1$ deg and $\sigma_{p} = 10$ cm, respectively.

\paragraph{\textbf{Performance}} We pose the visual SLAM system as a non-linear least squares problem, solved via Gaussian-Newton. Maximum 10 iterations are allowed for each method in this simulation for a fair comparison. We run the simulation sequence 30 times and show median results for the accuracy of the estimated trajectory and optimization time.
Fig.~\ref{fig:TIME_RMSE} shows similar performance across sequences, that is, $(-w)$ is more accurate and efficient than $(-r)$ and $(-wo)$. The second sequence (b) requires more optimization time and results in lower RMSE since more features are measured by each camera compared to the first. $P(-wo)$ requires less time than $P(-r)$ in two sequences because it does not use structural regularities and has small optimization computation as shown in Fig.~\ref{fig:parametrization1}(d).
$P(-r)$ has a higher computational burden (Fig.~\ref{fig:parametrization1}(e)) and is more accurate than $P(-wo)$. While combining line features in the system, like $PL(-r)$, results are more accurate even if the method requires more time. Compared to $P(-r)$ and $PL(-r)$, our parametrizations for points and lines ($P(-w)$) are more efficient. In terms of optimization time, $P(-w)$ has a 31\% improvement and $PL(-w)$ 33\%, compared to $P(-r)$.

\begin{table} 
\centering    
\caption{Computation time (mean, ms) of different operations in the V101 sequence of EuRoc. * means that the operation is used for each frame, otherwise it is performed on keyframes only. D\&M notes detection and matching. - means that the operation is not used. } 
\scalebox{0.8}{     
    \begin{tabular}{lcccccc}     \hline    
      Operation  &P(-wo) & P(-r) &P(-w) &PL(-wo) & PL(-r) &PL(-w) \\ \hline
    Point D\&M* &4 &4 &4 &4 &4 & 4 \\ 
    Line D\&M &- &- &- &96 &96 &96 \\ 
    Plane Seg. &- &29 &29 &- &29 &29 \\ \hline
    Plane fitting &- &10 &10 &- &10 &10  \\ 
Optimization &43 &44  &40 &36 &46 &42 \\  
Total time &56 &60 &55 &57 &58 &54  \\ \hline
\end{tabular}\label{table.real-time}
}
\end{table}
\begin{table}[h]  
\centering    
\caption{The number of landmarks updated in optimization module of sequence 2.} 
\scalebox{0.8}{     
    \begin{tabular}{ccccccc}     \hline    
        &P(-wo) & P(-r) &P(-w) &PL(-wo) & PL(-r) &PL(-w) \\ \hline     items &100 &101 &\textbf{51} & 120 &121 &\textbf{51} \\  parameters &350 &353 &\textbf{303} & 430 &433 &\textbf{303} \\ \hline
\end{tabular}\label{table. time2}
}
\end{table}
\paragraph{\textbf{Number of parameters}} 
Furthermore, we analyze the reason of efficiency from the perspective of number of parameters that are to be updated.
In traditional parametric methods (inverse depth for points and orthogonal approach for lines), each point, line and plane need 1 parameter, 4 parameters and 3 parameters, respectively. However, in our parametrization method, all points and lines in the plane are represented by only one plane parameter. Hence, there is only one parameter for each planar region during optimization. Tab.~\ref{table. time2} shows the number of parameters that need to be updated in the global bundle adjustment on the second Monte Carlo sequence, where 20 lines and 50 points are observed by 50 cameras. $P(-w)$ uses points only, so it has to update 100 items at each iteration. Similar to $P(-wo)$, we have to update 101 items and 121 items in $P(-r)$ and $PL(-r)$. Note that those two need to update one plane item because they use of co-planar constraints of point-to-plane and line-to-plane. In the proposed solutions, only 51 items (50 cameras and 1 plane) are updated in $P(-w)$ and $PL(-w)$ because they use the plane to represent co-planar points and lines. 

\section{CONCLUSION}
\label{sec:conc}

We presented an efficient and robust co-planar parametrization method for points and lines by leveraging geometric and learning approaches together, which increases sparsity and reduces the size of Hessian matrix in each optimization module. Then, we illustrated how our co-planar parametrization can be implemented in stereo-SLAM and VIO pipelines. Our experiments show that our approach improves the efficiency and accuracy of both stereo and VIO optimization in indoor environments. As for future work, we plan to reconstruct dense maps from monocular data and merge together semantic segmentation and depth prediction to improve tracking and mapping simultaneously. 




\ifCLASSOPTIONcaptionsoff
  \newpage
\fi

\end{document}